\documentclass[lettersize,journal]{IEEEtran}
\usepackage{amsmath,amsfonts}
\usepackage[normalem]{ulem}
\usepackage{color,soul}

\usepackage{graphics} 
\usepackage{epsfig} 
\usepackage{times} 
\usepackage{amsmath} 

\usepackage{amssymb}
\usepackage{yfonts}

\usepackage{algpseudocode}
\usepackage[ruled,vlined]{algorithm2e}

\usepackage{amsmath}

\usepackage{array}
\usepackage[caption=false]{subfig}
\usepackage{textcomp}
\usepackage{stfloats}
\usepackage{url}
\usepackage{verbatim}
\usepackage{graphicx}
\usepackage{cite}

\usepackage{booktabs}
\usepackage{footnote}
\usepackage{tabularx}
\usepackage{multirow}
\usepackage{hyperref}
\usepackage[flushleft]{threeparttable}
\usepackage{import}
\usepackage{siunitx}

\newcommand\T{{\hspace{-0pt}\intercal}}

\begin{document}
	
	\title{Ultrasound-Guided Assistive Robots for Scoliosis Assessment with Optimization-based Control and Variable Impedance}
	
	\author{Anqing Duan, Maria Victorova, Jingyuan Zhao, Yongping Zheng, and David Navarro-Alarcon%
	
	\thanks{This research is funded in part by the Research Impact Fund (RIF) of the HK Research Grants Council under grant R5017-18F, and in part by PolyU through the Intra-Faculty Interdisciplinary Project under grant ZVVR.}%
	
	\thanks{A Duan and D Navarro-Alarcon are with the Department of Mechanical Engineering, The Hong Kong Polytechnic University, KLN, Hong Kong.}
	\thanks{M Victorova, J Zhao and Y Zheng are with the Department of Biomedical Engineering, The Hong Kong Polytechnic University, KLN, Hong Kong.}}
	
	\markboth{}%
	{Duan \MakeLowercase{\textit{et al.}}: Assistive Robots for Scoliosis Assessment}
	\maketitle
	

	\begin{abstract}
		Assistive robots for healthcare have seen a growing demand due to the great potential of relieving medical practitioners from routine jobs. 
In this paper, we investigate the development of an optimization-based control framework for an ultrasound-guided assistive robot to perform scoliosis assessment. 
A conventional procedure for scoliosis assessment with ultrasound imaging typically requires a medical practitioner to slide an ultrasound probe along a patient's back.
To automate this type of procedure, we need to consider multiple objectives, such as contact force, position, orientation, energy, posture, etc.
To address the aforementioned components, we propose to formulate the control framework design as a quadratic programming problem with each objective weighed by its task priority subject to a set of equality and inequality constraints.
In addition, as the robot needs to establish constant contact with the patient during spine scanning, we incorporate variable impedance regulation of the end-effector position and orientation in the control architecture to enhance safety and stability during the physical human-robot interaction.
Wherein, the variable impedance gains are retrieved by learning from the medical expert's demonstrations.
The proposed methodology is evaluated by conducting real-world experiments of autonomous scoliosis assessment with a robot manipulator xArm.
The effectiveness is verified by the obtained coronal spinal images of both a phantom and a human subject.
	\end{abstract}
	
	
	\section{Introduction}\label{introduction}
	\IEEEPARstart{H}{ealthcare} and medical assistive robots have received increasing research attention due to growing demand in the market over the past decades \cite{okamura2010medical}.
As a typical modality of healthcare and medical assistive robot, ultrasound-guided navigator for physical body examination has numerous applications in clinical practice \cite{tirindelliral2020}.
Compared with X-rays, ultrasound has several advantages, such as no ionizing radiation and affordable cost.
Thus, it is favored in a wide spectrum of areas such as cardiology, urology, gynecology, etc \cite{li2021overview}.

\begin{figure}[t]
	\centering
	\includegraphics[width=0.9\columnwidth]{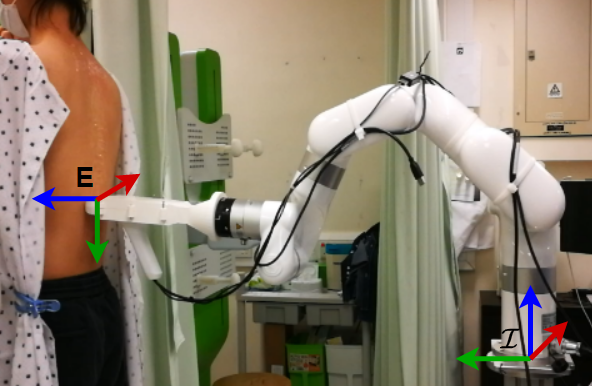}
	\caption{Illustration of the experimental setup with end-effector frame $E$ and base coordinate frame $\mathcal{I}$ labeled where $x, y, z-$axes are depicted in red, green, and blue, respectively. The Xarm manipulator is sliding an ultrasound probe along a human subject's back for scoliosis assessment.}
	\label{fig:exphuman}
\end{figure}

\begin{figure}[t]
	\centering
	\includegraphics[width=1\columnwidth]{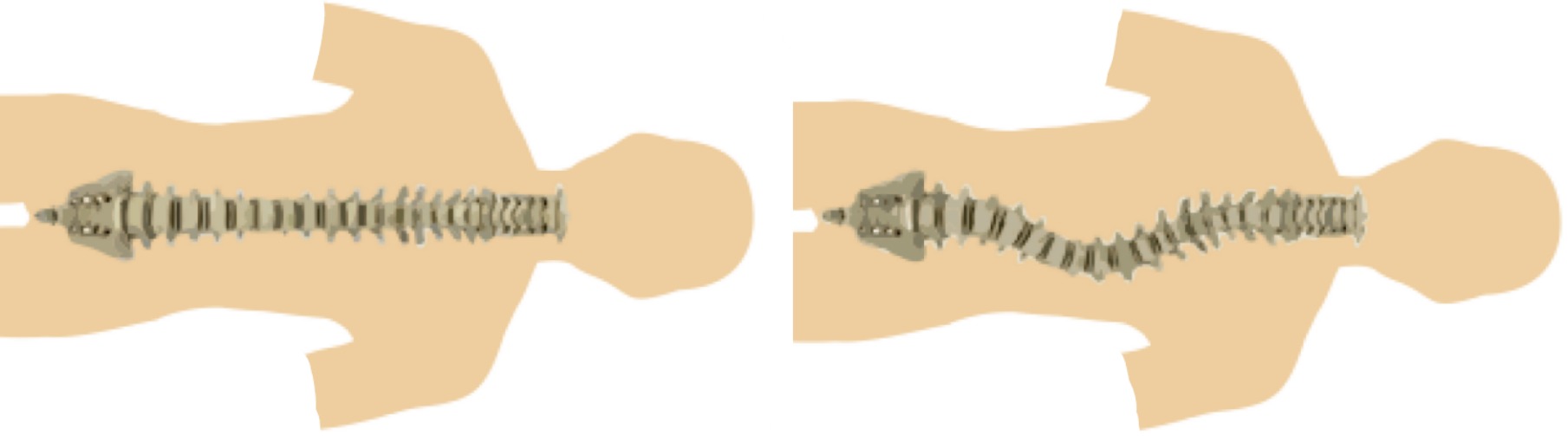}
	\caption{Comparison between the normal straight spine (\textit{left}) and the spine with scoliosis (\textit{right}).}
	\label{fig:scoliosis}
\end{figure}

\begin{figure}[t]
	\centering
	\includegraphics[width=1\columnwidth]{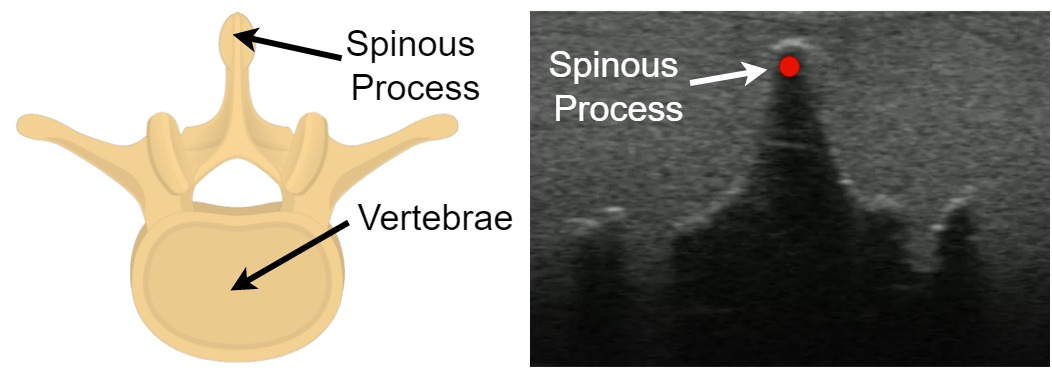}
	\caption{Anatomy illustration of vertebrae and its spinous process (\textit{left}) as well as the corresponding appearance in ultrasound image (\textit{right}) where the red dot denotes the detected spinous process.}
	\label{fig:verte_spin}
\end{figure}

In this paper, our focus lies on automating an ultrasound-guided robotic navigator for \textit{scoliosis assessment} as shown in Fig. \ref{fig:exphuman}. 
Scoliosis assessment is a type of physical examination needed for scoliosis progression screening, as shown in Fig. \ref{fig:scoliosis}, for timely spine correction and intervention.
To minimize the number of traditional X-ray scans needed and the exposure to ionizing radiation, a 3D spinal ultrasound is used for progression monitoring \cite{zheng2016reliability}. 
The process of image generation usually requires the human operator to carefully scan the spine's profile by sliding an ultrasound probe along the patient's back.
To ensure that the quality of the ultrasound image is satisfactory, the operator needs to adjust the pose of the probe in real-time during the scan procedure such that the vertebrae (whose presence is normally indicated by the spinous process) are located in the middle of the image, as illustrated in Fig. \ref{fig:verte_spin}.
Therefore, the key to the success of 3D spine image reconstruction is to locate and follow the spinous process closely while sweeping the probe along the spine's curvature. 
The coronal slice of the resulting 3D reconstruction is used for the lateral scoliosis curvature measurement. 
From a robotic point of view, we identify that automating scoliosis assessment heavily relies on technical support from two main research areas, namely, ultrasound-guided navigation \cite{hase2020ultrasound} and physical human-robot interaction (pHRI) with \textit{variable impedance control} \cite{abu2020variable}.

Ultrasound-based strategies for navigation have been applied to a variety of medical devices and surgical instruments due to their minimally invasive intervention properties and high portability.
Until recently, several attempts towards medical procedures through an ultrasound-guided robot manipulator have emerged.  
It is observed, however, that the majority of solutions to the probe's motion generation are neural network-based controllers whose parameters are trained either by reinforcement learning \cite{hase2020ultrasound, li2021image} or human demonstrations \cite{deng2021learning}.
Yet, data efficiency issues, as well as explainability of the learning-based controllers, draw a big question mark under the context of medical applications.

The formalism of our application, a hybrid motion/force control on a human body, is also analyzed to fall into a category of pHRI.
Notably, a rich literature on pHRI is dedicated to enhancing safety and flexibility \cite{hu2020interact}.
Among various perspectives, we aim at endowing the robot with variable impedance control, which plays a crucial role for a robot to safely perform contact-sensitive tasks \cite{boaventura2015model, huokeypoint} or reliably assist a human subject for rehabilitation purposes \cite{lee2020modulating}. 
Compared with stochastic searching-based techniques such as evolution strategies \cite{hu2018evolution} or inverse reinforcement learning \cite{zhang2021learning}, transferring variable impedance skills via human demonstrations provides a more intuitive and straightforward fashion \cite{yang2018dmps, duan2019learning}.
In addition, most existing works on impedance regulation revolve around either robot joint space or end-effector position \cite{duan2020learning}, while it is usually overlooked to incorporate impedance regulation for orientation.
A comparison with state-of-the-art methods of robotic ultrasound scanning is shown in Table \ref{Table:comp}.
 
Previously, we devised a novel algorithm for trajectory planning based on spinous process localization in ultrasound images \cite{victorova2021follow}.
Although the preliminary results have shown that a robotic approach is promising for spinal image reconstruction, there is a lack of a principled control theoretic framework for prioritizing different tasks, such as contact force, end-effector pose, robot configuration, etc, under external contact constraints.
In view of that, in this paper, we propose a new optimization-based control architecture capable of considering multiple objectives simultaneously while respecting a set of equality and inequality constraints.
Similar to \cite{nava2016stability}, our controller has two loops and is implemented via quadratic programming (QP), which has shown superior performance in many safety-critical applications, such as humanoid locomotion \cite{nava2016stability}, aerial manipulation \cite{nava2019direct}, and flying humanoid \cite{pucci2017momentum}.
Furthermore, regarding acquisition of variable impedance gains, we follow the general paradigm of learning by demonstration \cite{calinon2016tutorial}.
Specifically, to incorporate the impedance regulation of the probe's orientation, we propose to parameterize the demonstrated rotation matrix trajectory such that the issue of exploiting the covariance matrix for a variable in a form other than a vector can be alleviated.


To the best of the authors' knowledge, this is the first time that the proposed methodology has been used to automate the probe's manipulation during scoliosis assessment.
The main contributions of this work can be summarized as follows:
\begin{itemize}
    \item Development of a new optimization-based control architecture for autonomous scoliosis assessment;
    \item Incorporation of impedance regulation via learning by demonstrations of the ultrasound probe's manipulation;
    \item Experimental validation of the proposed theory with spine phantom models and human subjects\footnote{Ethical approval HSEARS20210417002 was given by the Departmental
Research Committee on behalf of PolyU Institutional Review Board.}.
\end{itemize}

\begin{table}[t]
\centering
\caption{Comparison with State-of-the-art Approaches.}
\begin{tabular}[t]{lcccc}
\toprule
&\begin{tabular}{@{}c@{}}Contact \\ Awareness\end{tabular} & \begin{tabular}{@{}c@{}} Control \\ Constraints\end{tabular} & \begin{tabular}{@{}c@{}}Variable \\ Impedance\end{tabular} & \begin{tabular}{@{}c@{}}From \\ Demonstrations\end{tabular}\\
\midrule
\cite{tirindelliral2020}& \checkmark & -- &--& --\\
\cite{li2021image} &--&\checkmark &--& --\\
\cite{deng2021learning} &\checkmark&-- & -- & \checkmark\\
\cite{victorova2021follow}&\checkmark &-- &-- & --\\
Ours & 	\checkmark & 	\checkmark & 	\checkmark & 	\checkmark \\
\bottomrule
\end{tabular}
\label{Table:comp}
\end{table}
The rest of the paper is organized as follows:
Section \ref{control} presents the mathematical models; Section \ref{learnimp} presents the controller design; Section \ref{experiments} describes the results; Section \ref{conclusion} presents discussions and gives final conclusions.

	\section{Control Architecture Design}\label{control}

\subsection{Dynamics Modeling}\label{dynmodel}
Recall that the joint space dynamics equation of motion for a fixed-base and open-chain robot manipulator whose configuration is characterized by the joint angles $\mathbf{q} \in \mathbb{R}^n$ can be modeled as \cite{dna2014tcst}:
\begin{equation}\label{robotsys}
\mathbf{M}(\mathbf{q})\ddot{\mathbf{q}}+\mathbf{c}(\mathbf{q}, \dot{\mathbf{q}}) + \boldsymbol{g}(\mathbf{q}) = \boldsymbol{\tau} + \mathbf{J}(\mathbf{q})^\T\boldsymbol{f},
\end{equation}
where $\mathbf{M}(\mathbf{q}) \in \mathbb{R}^{n\times n}$ is the mass matrix, $\mathbf{c}(\mathbf{q}, \dot{\mathbf{q}}) \in \mathbb{R}^{n}$ accounts for the Coriolis and centrifugal effects, $\boldsymbol{g}(\mathbf{q}) \in \mathbb{R}^{n}$ is the gravity vector, $\boldsymbol{\tau}\in \mathbb{R}^{n}$ are the joint actuation torques.
When an external wrench $\boldsymbol{f}$, which is expressed in a frame that has the same orientation as the world frame $\mathcal{I}$, is exerted on the robot manipulator, its effects on the robot dynamics are reflected by the Jacobian matrix $\mathbf{J}(\mathbf{q}) \in \mathbb{R}^{6\times n}$ that maps the robot joints velocity $\dot{\mathbf{q}}$ to the linear and angular velocities of the contact point where $\boldsymbol{f}$ is applied.


When the robot sweeps the ultrasound probe along the subject's back, it exhibits a hybrid motion/force behavior.
We consider to model the interaction between the robot end-effector and the human back with a set of holonomic constraints.
By considering the constraints of the end-effector motion explicitly, the robot shall possess higher safety level during the diagnostic procedure.
Specifically, the time derivatives of these constraints are expressed as
\begin{equation}\label{constraints}
\mathbf{B}_c^\T [\mathbf{v}_b^\T \;~ \boldsymbol{\omega}_b^\T]^\T = \mathbf{G}^\T[\mathbf{v}_E^\T \;~ \boldsymbol{\omega}_E^\T]^\T=
 \mathbf{J}_c\dot{\mathbf{q}} = \mathbf{0}_{n_c},
\end{equation}
where we define $\mathbf{J}_c = \mathbf{G}^\T\mathbf{J}$ with $\mathbf{G} = \mathbf{Ad}_{g^{-1}}^\T\mathbf{B}_c \in \mathbb{R}^{6\times n_c}$ being the contact map and $\mathbf{Ad}_{g} = \mathtt{blkdiag}(\!\!\!~^\mathcal{I}\mathbf{R}_E, ~^\mathcal{I}\mathbf{R}_E)$ being the adjoint transformation matrix\footnote{As an abuse of notation, it is not exactly the mapping from body velocity to spatial velocity.} that maps robot end-effector \textit{body} velocity to the linear and angular velocities expressed in the world frame $\mathcal{I}$ with $~^\mathcal{I}\mathbf{R}_E \in SO(3)$ being the rotation matrix expressing the orientation of the end-effector frame $E$ with respect to $\mathcal{I}$. $\mathbf{v}_b$ and $\boldsymbol{\omega}_b$ are the linear and angular body velocities that are expressed in the instantaneous body frame.
$\mathbf{B}_c \in \mathbb{R}^{6\times n_c}$ is the wrench basis with $n_c$ indicating the number of constrained directions of motion (or equivalently, independent forces) of the robot end-effector. 
$\boldsymbol{\omega}_E$ is the spatial angular velocity of the end-effector such that $\dot{\mathbf{R}}_E=S(\boldsymbol{\omega}_E)\mathbf{R}_E$ where $S$ is the skew operator\footnote{Reference frame $\mathcal{I}$ is omitted for brevity from now on.}. 

Likewise, the equation expressing the unconstrained direction of motion $\mathbf{v}_u$ can be written as
\begin{equation}\label{unconstrainedmotion}
\mathbf{B}_u^{\T}[\mathbf{v}_b^\T \;~ \boldsymbol{\omega}_b^\T]^\T = \mathbf{G}_u^\T[\mathbf{v}_E^\T \;~ \boldsymbol{\omega}_E^\T] = \mathbf{J}_u\dot{\mathbf{q}} = \mathbf{v}_u,
\end{equation}
where $\mathbf{v}_u$ denotes the velocity of the unconstrained direction of motion expressed in the body frame and we define $\mathbf{J}_u = \mathbf{G}_u^\T\mathbf{J}$ with $\mathbf{G}_u = \mathbf{Ad}_{g^{-1}}^\T\mathbf{B}_u$ and $\mathbf{B}_u \in \mathbb{R}^{6\times 6-n_c}$ being complementary to $\mathbf{B}_c$. 
We then rewrite the robot dynamics equation \eqref{robotsys} by additionally including the contact constraints and decoupling the external wrench $\boldsymbol{f}$ as a result of non-holonomic constraints as well as free motion\footnote{Dependence on the robot states is dropped for brevity from now on.}:
\begin{subequations}\label{constraintdyn}
\begin{equation}\label{sysdec}
\mathbf{M}\ddot{\mathbf{q}}+\mathbf{c} + \boldsymbol{g} = \boldsymbol{\tau} + \mathbf{J}_c^\T\boldsymbol{f}_c +  \mathbf{J}_u^\T\boldsymbol{f}_u,
\end{equation}
\begin{equation}\label{diffcon}
\mathbf{J}_c\dot{\mathbf{q}} + \dot{\mathbf{J}}_c\ddot{\mathbf{q}} = \mathbf{0},	
\end{equation}
\end{subequations}
where $\boldsymbol{f}_c \in \mathbb{R}^{n_c}$ denote the forces due to the existence of the constraints and $\boldsymbol{f}_u \in \mathbb{R}^{6-n_c}$ represent the forces that are caused by allowable directions of motion such as frictions. 
The constraints on the acceleration of the end-effector motion \eqref{diffcon} is obtained by differentiating \eqref{constraints} with respect to time.


\begin{figure}[t]
	\centering
	\includegraphics[width=1\columnwidth]{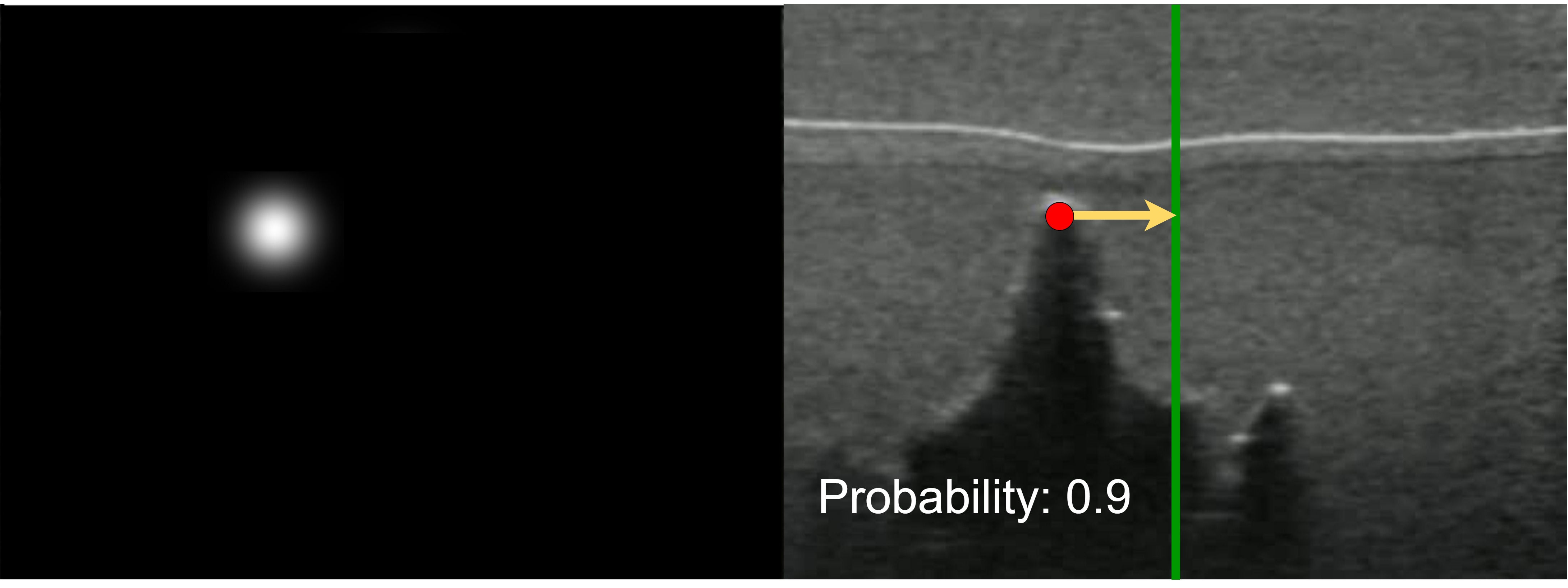}
	\caption{Illustration of the adjustment strategy for the ultrasound probe. From the heatmap (\textit{left}) of the original ultrasound image of the spinous process (\textit{right}), the deviation (yellow arrow) away from the middle line (green) is used to laterally guide the probe.}
	\label{fig:HeatMap}
\end{figure}

\subsection{Task Specifications}\label{tasksspec}
In order to make the robot automatically perform scoliosis assessment, it is necessary to properly determine a vector of quantities of interest $\boldsymbol{\zeta}$ such that by tracking the corresponding desired values denoted by $\boldsymbol{\zeta}^d$, the robot is able to function and achieve the goal. 
Wherein, each term of the output vector $\boldsymbol{\zeta}$ is called a \textit{task} in the language of robot control.
The selected tasks that are considered to be relevant to our application are listed as follows:  
\begin{itemize}
\item Joints configuration of the robotic manipulator $\mathbf{q}$;
\item Linear position due to the unconstrained motion $\mathbf{p}_u$;
\item Angular position due to the unconstrained motion $\mathbf{R}_E$; 
\item Contact forces arising from the constraints $\boldsymbol{f}_c$.
\end{itemize}
The tasks chosen to control are collectively expressed as
\begin{equation}
\boldsymbol{\zeta} = [\mathbf{q}^\T \;~ \mathbf{p}_u^\T \;~ \mathbf{R}_E^\T \;~ \boldsymbol{f}_c^\T]^\T.    
\end{equation}

The proper design of $\boldsymbol{\zeta}^d$ plays an important role in achieving scoliosis assessment.   
The key to the success of the spinal image reconstruction lies on fine-tuning of the $x$-direction movement of the body frame. 
To this end, we previously developed a novel fully connected network built upon ResNet that is suitable for spinous localization. 
The proposed in \cite{victorova2021follow} neural network takes as input the raw ultrasound
image and outputs a spatial heatmap that indicates the location of the spinous process as well as the confidence probability.
The distance between the detected spinous location and the image
center is then used to guide the movement of $x$-direction of the body frame such that the ultrasound probe will move in a way that the vertebrae is always kept in the center of the ultrasound image, as shown in Fig. \ref{fig:HeatMap}.
Furthermore, in order to drive the ultrasound probe moving along the patient's spine from the waist to the neck, $y$-direction of the body frame of the ultrasound probe is empirically set to be a constant velocity.
For the probe's orientation control, the desired rotation matrix is determined such that the probe is normally pointing towards the human back surface.
The reference force profile is empirically set based on the subject's body mass index to ensure tight contact with skin \cite{victorova2021follow}.

\subsection{Controller Design}\label{controldesign}
Next, we consider the design of a controller for dynamical system \eqref{constraintdyn} to achieve the aforementioned tasks.
The goal here is to make the quantities of interest $\boldsymbol{\zeta}$ track the desired trajectory $\boldsymbol{\zeta}^d$ that is specified by our ultrasound image processor.
In view of high complexity of the multi-input multi-output system in addition to several constraints that could emerge, such as control bounds, joint limits, kinematic constraints etc., we consider to formulate our control problem from an optimization perspective.
Compared with analytic control law design, optimization-based control strategies exhibit great potential for customization towards different requirements \cite{li2016geometric} and better at explicitly handling constraints \cite{duan2018constrained}.
Specifically, our control method akin to a feedback linearization method is composed of two loops similar to \cite{nava2019direct}.
In the outer loop, it is assumed that higher-order derivatives of tasks $\boldsymbol{\zeta}$ defined as 
\begin{equation}
\mathbf{v} = [\ddot{\mathbf{q}}^\T \;~ \ddot{\mathbf{p}}_u^\T\;~ \dot{\boldsymbol{\omega}}_E^\T\;~ \boldsymbol{f}_c^\T]^\T    
\end{equation}
is directly controllable by a so-called virtual input $\mathbf{v}^*$. 
The input-output asymptotic stability is guaranteed if there exists a control signal $\mathbf{u}^*$ such that $\mathbf{v}^*$ can be exactly achieved, i.e. $\mathbf{v} = \mathbf{v}^*$.
In principle, any forms stabilizing controller that can make $\boldsymbol{\zeta}$ smoothly track $\boldsymbol{\zeta}^d$ is valid. 

Here, the chosen stabilizing controller is designed in the form of a PID controller due to its simplicity and the virtual input is designed as:
\begin{equation}\label{outterloop}
	\mathbf{v}^* = \begin{bmatrix}
		\ddot{\mathbf{q}}^d- \mathbf{K}_{q}^D(\dot{\mathbf{q}}-\dot{\mathbf{q}}^d) -\mathbf{K}_{q}^P(\mathbf{q}-\mathbf{q}^d)  \\
		\ddot{\mathbf{p}}_u^d- \mathbf{K}_{u}^D(\dot{\mathbf{p}}_{u}-\dot{\mathbf{p}}_{u}^d) -\mathbf{K}_{u}^P(\mathbf{p}_{u}-\mathbf{p}_{u}^d)\\
		\dot{\boldsymbol{\omega}}_E^d - \mathbf{K}_{R}^D(\boldsymbol{\omega}_E-\boldsymbol{\omega}_E^d) - \mathbf{K}_{R}^P\mathbf{e}_R   \\
		\boldsymbol{f}_c^d- \mathbf{K}^P_c (\boldsymbol{f}_c-\boldsymbol{f}_c^d) - \mathbf{K}^I_c\int_{0}^{t} (\boldsymbol{f}_c-\boldsymbol{f}_c^d) dt
	\end{bmatrix}.
\end{equation}
Wherein, the rotation error $\mathbf{e}_R$ between two rotation matrices: the desired rotation matrix $\mathbf{R}_E^d$ and current end-effector rotation matrix $\mathbf{R}_E$ is defined as \cite{zhu2017some}
\begin{equation}\label{roterr}
	\mathbf{e}_R = \frac{1}{2}(\mathbf{R}_E^\T\mathbf{R}_E^d- \mathbf{R}_E^{d\T}\mathbf{R}_E)^\vee \in \mathbb{R}^3,
\end{equation}
where $(\cdot)^\vee$ denotes the inverse of the skew operator.



For the inner loop, it is responsible to generate control signals $\boldsymbol{\tau}$ for the tracking of the virtual input $\mathbf{v}^*$.
Typically, it is usually very difficult to design an analytical control law for a multi-input (namely $\boldsymbol{\tau}$) multi-output (namely $\mathbf{v}$) system, it is thus considered to leverage optimization-based techniques as they require less manual efforts compared with analytical control law design.
To do so, we first write the system dynamics equation of the inner loop in a control-affine style as follows
\begin{equation}
\mathbf{v} = \mathbf{A}(\mathbf{q})\mathbf{u} + \mathbf{b}(\mathbf{q}, \dot{\mathbf{q}}), \quad \text{with} \quad \mathbf{u} = [\boldsymbol{\tau}^\T \;~ \boldsymbol{f}_c^\T]^\T.
\end{equation}
It should be noted here that contact force $\boldsymbol{f}_c$ is also included as a part of the control signals $\mathbf{u}$ in addition to joint torques $\boldsymbol{\tau}$, which implies that 
it is both a task vector and a control input.
And thus the order of its derivative in $\mathbf{v}$ is zero.
In the system dynamics, the matrix $\mathbf{A}(\mathbf{q})$ and the bias vector $\mathbf{b}(\mathbf{q}, \dot{\mathbf{q}})$ that contain all the terms independent of the extended control input $\mathbf{u}$ are given by
\begin{equation}
\mathbf{A} = \begin{bmatrix}
		\mathbf{M}^{-1} & \mathbf{M}^{-1}\mathbf{J}_c^\T \\
		\mathbf{J}_u\mathbf{M}^{-1} &  \mathbf{J}_u\mathbf{M}^{-1}\mathbf{J}_c^\T\\
		\mathbf{0}_{n_c\times n} &  \mathbf{I}_{n_c}
\end{bmatrix}, \quad
\mathbf{b} = \begin{bmatrix}
	\boldsymbol{\Gamma} \\
	\mathbf{J}_u\boldsymbol{\Gamma} + \dot{\mathbf{J}}_u\dot{\mathbf{q}}\\
	\mathbf{0}_{n_c}
\end{bmatrix},
\end{equation}
where we denote $\boldsymbol{\Gamma} = \mathbf{M}^{-1}(\mathbf{J}_u^\T\boldsymbol{f}_u-\mathbf{c}-\boldsymbol{g})$.
The optimization problem to be solved in the inner loop is formulated as:
\begin{subequations}\label{inneropt}
\begin{align}
& \,\quad\quad \underset{\mathbf{u}}{\min} 
&  \|\mathbf{v}-&\mathbf{v}^*\|_{\mathbf{W}}^2 & \label{ctrlobj}\\
&\;\,\quad\quad\mathtt{s.t.} & \underline{\mathbf{u}} \leq &\mathbf{u} \leq \overline{\mathbf{u}} \label{controlbound}\\
&& \mathbf{C}\boldsymbol{f}&_c\leq\mathbf{0} \label{friction}\\
&& \mathbf{M}\ddot{\mathbf{q}}+\mathbf{c} + \boldsymbol{g} = &\boldsymbol{\tau} + \mathbf{J}_c^\T\boldsymbol{f}_c +  \mathbf{J}_u^\T\boldsymbol{f}_u \label{sysdyn}\\
&& \mathbf{J}_c\dot{\mathbf{q}} + &\dot{\mathbf{J}}_c\ddot{\mathbf{q}} = \mathbf{0}\label{kincon}
\end{align}
\end{subequations}
The control objective \eqref{ctrlobj} is designed as a squared weighted Euclidean norm $\|\cdot\|_{\mathbf{W}}^2$ where each task priorities can be specified by $\mathbf{W}$.
And the constraint \eqref{controlbound} specifies the bounds on the control input such as joint torques limits and contact force magnitude, which are critical when establishing interactions with humans for medical purposes.
The specified upper bound and lower bound are denoted by $\overline{\mathbf{u}}$ and $\underline{\mathbf{u}}$, respectively.
The constraint \eqref{friction} requests that the contact friction cones $FC$ are respected by approximating $\boldsymbol{f}_c \in FC$ with linear inequality.
\eqref{sysdyn} impose constraints due to system dynamics while \eqref{kincon} correspond to the kinematic constraints as a result of contact with the environment.

It can be verified that \eqref{inneropt} in fact belongs to a Quadratic Programming (QP) problem.
In general, the standard formulation of a QP problem is composed of a quadratic-form objective function of the design variables and a set of linear equality and inequality constraints:
\begin{equation}\label{generalqp}
\begin{aligned}
\underset{\mathbf{u}}{\min}\;\frac{1}{2}\mathbf{u}^\T\mathbf{P}_c\mathbf{u}+\mathbf{q}_c^\T\mathbf{u} 
\quad \mathtt{s.t.}
& & \mathbf{L}\mathbf{u} \leq \mathbf{h} \;\text{and}\; \mathbf{D}\mathbf{u} = \mathbf{z}.
\end{aligned}
\end{equation}

\begin{figure}[t]
	\centering
	\includegraphics[width=1\columnwidth]{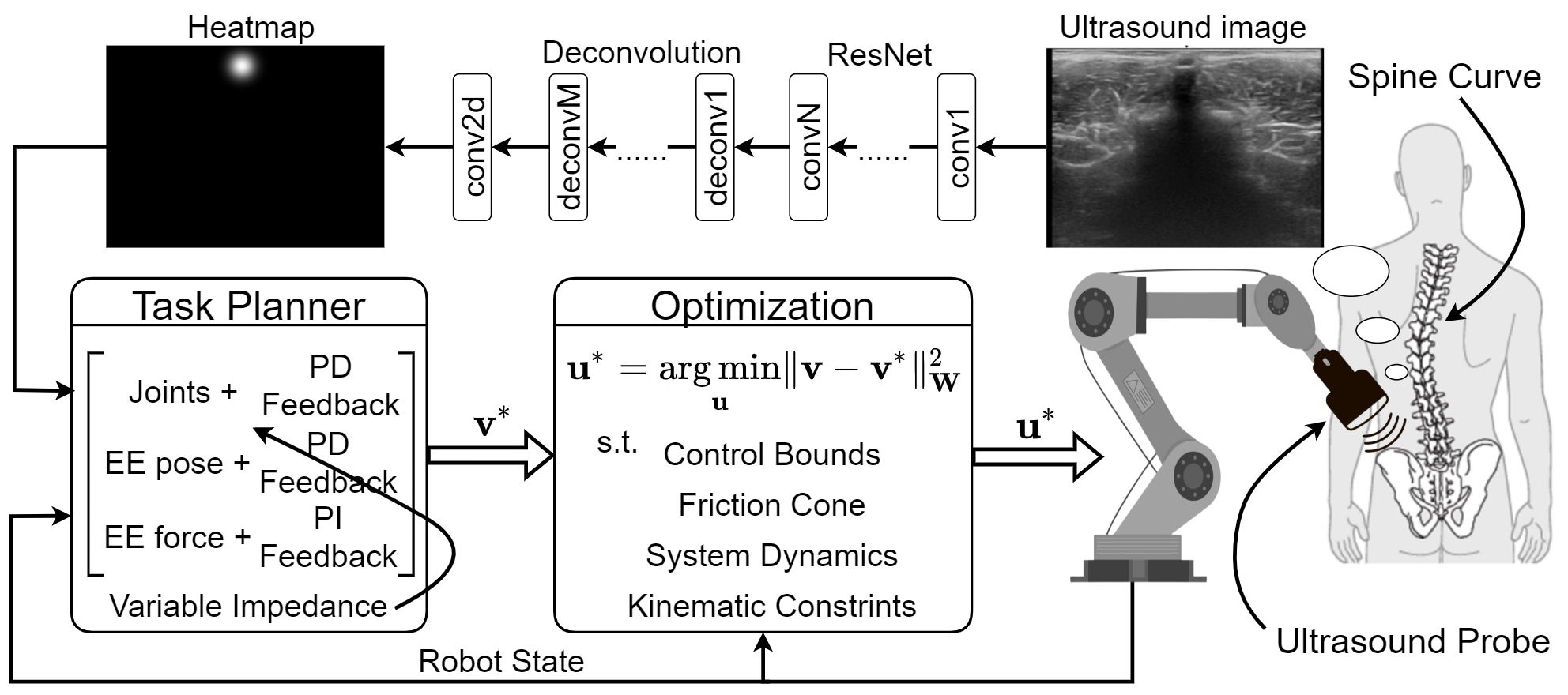}
	\caption{Illustration of the overall proposed control architecture for autonomous scoliosis assessment. First, the ultrasound images are processed with a fully connected network, which outputs the predicted location of the spinous process. Then the outer loop of the controller sends higher-order derivatives of the desired values of quantities of interest while the inner loop solves an optimization problem for robot control commands.
}
	\label{fig:motionillu}
\end{figure}

By re-arranging the objectives and constraint terms of \eqref{inneropt} towards the general QP formalism of \eqref{generalqp}, the original optimization problem \eqref{inneropt} can be equivalently written as
\begin{equation}\label{qpreform}
\begin{aligned}
\min_{\mathbf{u}} \quad  \frac{1}{2}\mathbf{u}^\T\mathbf{A}^\T\mathbf{W}\mathbf{A}\mathbf{u}+&(\mathbf{b} - \mathbf{v}^*)^\T\mathbf{W}\mathbf{A}\mathbf{u}\\
\mathtt{s.t.\quad\,\;} \quad  \begin{bmatrix} \mathbf{I}_{n+n_c}\\
-\mathbf{I}_{n+n_c}\\
\mathbf{0}_{n}^\T \; \mathbf{C}
\end{bmatrix}\mathbf{u} &\leq \begin{bmatrix} \overline{\mathbf{u}}\\
-\underline{\mathbf{u}}\\
\mathbf{0}
\end{bmatrix}\\
\begin{bmatrix}\dot{\mathbf{J}}_c\mathbf{M}^{-1} \;~ \dot{\mathbf{J}}_c\mathbf{M}^{-1}\mathbf{J}_c^\T  \end{bmatrix}&\mathbf{u} = -\mathbf{J}_c\dot{\mathbf{q}}- \dot{\mathbf{J}}_c\boldsymbol{\Gamma}.\\
\end{aligned}
\end{equation}
The correspondence relationship between \eqref{inneropt} and \eqref{generalqp} is thus yielded explicitly with the help of \eqref{qpreform}.
As a result, the inner-loop optimization can be solved fast enough due to the merits of QP and it is possible to implement real-time optimization-based robot control.  
A diagram of the overall robotic system for scoliosis assessment is illustrated in Fig. \ref{fig:motionillu}. 

	\section{Impedance Gain Retrieval}\label{learnimp}
\begin{figure}[t]
	\centering
	\includegraphics[width=0.9\columnwidth]{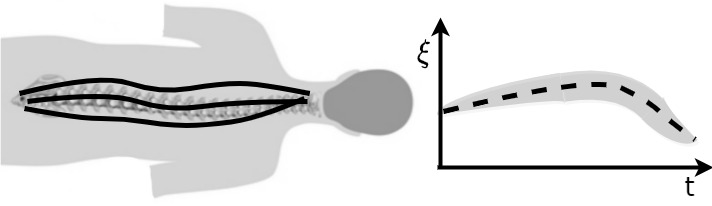}
	\caption{Illustration of the learning by demonstration approach for the retrieval of impedance gains. The desired impedance gain is obtained by estimation of covariance matrix from multiple demonstrations.}
	\label{fig:motion}
\end{figure}

	In this section, we study the problem of regulating the impedance gains, namely the stiffness and damping matrices $\mathbf{K}_u^D$, $\mathbf{K}_u^P$, $\mathbf{K}_R^D$, and $\mathbf{K}_R^P$ of \eqref{outterloop}, to guarantee safe and reliable interaction with the human patients.
To this end, we consider to retrieve the desired impedance gains by resorting to the general framework of learning by demonstration or imitation learning since it provides a straightforward and intuitive manner to transfer human motion skills to robots \cite{Omar2022snn}.
Specifically, we propose to leverage the techniques from probabilistic imitation learning framework, where we can obtain variability from multiple human demonstrations to regulate the robot manipulator's impedance as shown in Fig. \ref{fig:motion}.
To retrieve the desired impedance profile from the demonstrated trajectories, the region where the dispersion of the trajectories is high implies that robots will lower stiffness gains. 

One popular approach is to encode variability of the demonstrated trajectories through Gaussian Mixture Regression (GMR) \cite{calinon2016tutorial}.
Without loss of generality, we denote $\boldsymbol{\xi}$ to represent either one of $\mathbf{p}_u$, $\dot{\mathbf{p}}_u$, and $\boldsymbol{\omega}_E$ whereas the retrieval of the orientation stiffness profile will be explained later as the rotation matrix does not follow the Euclidean distance metric.

Given $M$ demonstrations from an expert, we can collect a dataset $\{\{d_n^m, \boldsymbol{\xi}_n^m\}_{n=1}^{N}\}_{m=1}^{M}$ with each demonstration length being $N$. 
The input $d \in \mathbb{R}$ is chosen as time stamps. 
For the purpose of generalization or different test speed, we could simply scale it accordingly.
  
In order to retrieve the variability of the demonstrated quantities with GMR from the dataset, we first need to encode the joint probability distribution of input $d$ and output $\boldsymbol{\xi}$ with the Gaussian Mixture Model (GMM) representation: $\mathcal{P}(d, \boldsymbol{\xi}) = \sum_{k=1}^K\pi_k\mathcal{N}(\boldsymbol{\mu}_k, \boldsymbol{\Sigma}_k)$ where $\pi_k$ denotes the prior probability of each Gaussian component in total of $K$ Gaussian components with $ \sum_{k=1}^K\pi_k = 1$, and 
\begin{equation}\label{GMMmeancov}
\boldsymbol{\mu}_k= \begin{bmatrix}
	\boldsymbol{\mu}_{d,k}\\
	\boldsymbol{\mu}_{\xi,k}
\end{bmatrix}\quad \text{and} \quad
\boldsymbol{\Sigma}_k= \begin{bmatrix}
	\boldsymbol{\Sigma}_{dd,k} & \boldsymbol{\Sigma}_{d\xi,k}\\
	\boldsymbol{\Sigma}_{\xi d,k} &\boldsymbol{\Sigma}_{\xi\xi,k} 
\end{bmatrix}.
\end{equation}
During the reproduction phase, the conditional distribution $\mathcal{P}(\boldsymbol{\xi}(d))$ given a query point $d$ is given as  
\begin{equation}\label{multimodal}
\mathcal{P}(\boldsymbol{\xi}(d)) = \sum_{k=1}^{K}\eta_k(d)\mathcal{N}(\boldsymbol{\mu}_{k|d}, \boldsymbol{\Sigma}_{k|d}),
\end{equation}
where the weighting term $\eta_k(d)$, the conditional mean $\boldsymbol{\mu}_{k|d}$ and the covariance of a Gaussian component $\boldsymbol{\Sigma}_{k|d}$ are respectively given by \cite{calinon2016tutorial}
\begin{subequations}
	\begin{equation}
		\eta_k(d) = \frac{\pi_k\mathcal{N}(\boldsymbol{\mu}_{d,k}, \boldsymbol{\Sigma}_{dd,k})}{\sum_{j=1}^K\pi_j \mathcal{N}(\boldsymbol{\mu}_{d,j}, \boldsymbol{\Sigma}_{dd,j})},
	\end{equation}
\begin{equation}
	\boldsymbol{\mu}_{k|d} = \boldsymbol{\mu}_{\xi,k}+  \boldsymbol{\Sigma}_{\xi d,k} \boldsymbol{\Sigma}_{d d,k}^{-1}(d-\boldsymbol{\mu}_{d,k}),                 
\end{equation}
\begin{equation}
	\boldsymbol{\Sigma}_{k|d} = \boldsymbol{\Sigma}_{\xi\xi,k}-\boldsymbol{\Sigma}_{\xi d,k}\boldsymbol{\Sigma}_{dd,k}^{-1}\boldsymbol{\Sigma}_{d\xi,k}.    
\end{equation}
\end{subequations}

Notably, we are especially interested in retrieving the covariance matrix whose inverse serves as a proxy of the impedance gains.
Therefore, the single peaked covariance of the multi-modal distribution \eqref{multimodal} is approximated with
\begin{equation}\label{covmat}
	\mathtt{cov}(\boldsymbol{\xi}(d)) = \sum_{k=1}^{K}\eta_k(d)(\boldsymbol{\Sigma}_{k|d}+\boldsymbol{\mu}_{k|d}\boldsymbol{\mu}_{k|d}^\T) - \boldsymbol{\mu}_d\boldsymbol{\mu}_d^\T,
\end{equation}
where the conditional mean value of the approximated single normal distribution is calculated as
$\boldsymbol{\mu}_d = \sum_{k=1}^{K}\eta_k(d)\boldsymbol{\mu}_{k|d}$.
Finally, the impedance gains  $\mathbf{K}_u^D$, $\mathbf{K}_u^P$, and $\mathbf{K}_R^D$ are then set as the inverse of \eqref{covmat}.

It should be noted that special attention shall be paid to the calculation of the stiffness matrix $\mathbf{K}_R^P$ for rotation control. 
Indeed, the problem is not trivial, if not ill-posed, when directly applying GMM/GMR to quantifying the aleatoric uncertainty of a variable in the form of a matrix rather than a vector, as in our case of rotation matrix.
As a workaround, we propose to parameterize a rotation matrix with its exponential coordinate, namely $\mathbf{R} = \exp\left({S(\boldsymbol{\omega})\theta}\right)$ 
with $\boldsymbol{\omega} \in \mathbb{R}^3$ being a unit vector and $\theta\in \mathbb{R}$ given by \cite{murray2017mathematical}
\begin{equation}
	\theta = \cos^{-1} \left(\frac{\mathrm{tr}(\mathbf{R})-1}{2}\right), \quad
	\boldsymbol{\omega} = \frac{1}{2\sin(\theta)} (\mathbf{R}-\mathbf{R}^\T)^\vee.
\end{equation}
Consequently, with such parameterization of the end-effector's orientation, the rotational stiffness matrix $\mathbf{K}_R^P$ is retrieved by instantiating $\boldsymbol{\xi}$ with $\boldsymbol{\omega}\theta$ and then following a similar procedure as per \eqref{GMMmeancov}$-$\eqref{covmat}.
By doing so, we can avoid employing complicated manifold-based regression algorithms to deal with the manifold-structured rotation matrix data.
Arguably, the eligibility of employing the inverse of the covariance matrix of $\boldsymbol{\omega}\theta$ to represent $\mathbf{K}_R^P$ is evidenced by its compatibility with our choice of the rotation error as in \eqref{roterr}.



	\section{Results}\label{experiments}
\begin{figure}[t]
	\centering
	\includegraphics[width=0.9\columnwidth]{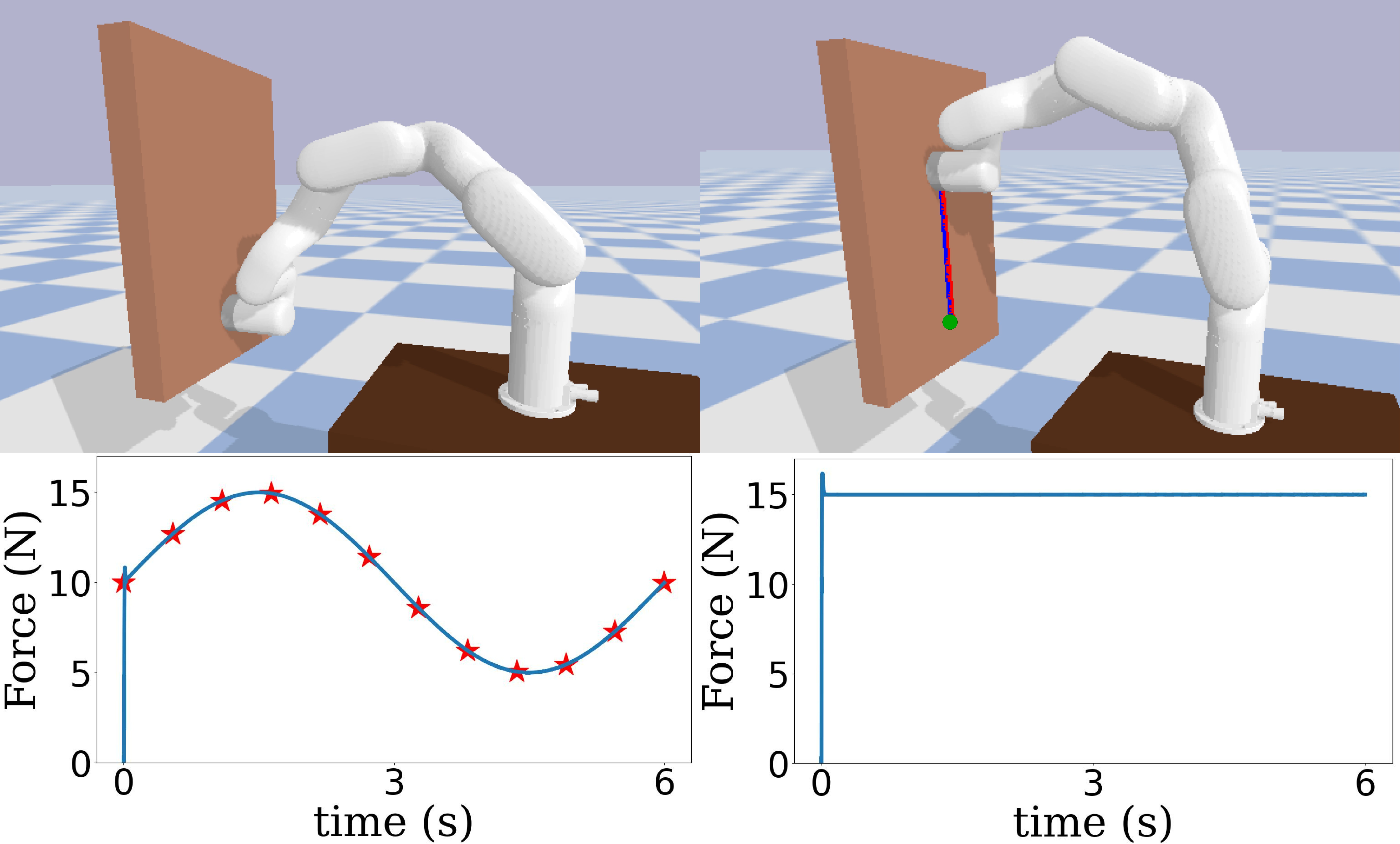}
	\caption{Snapshot of simulation. The robot in start position (\textit{top left}) and end position (\textit{top right}), where green dot denotes the start point, red line is reference of end-effector position, 
	and blue line are the real positions of the end-effector. The plots represent the force tracking performance with red stars denoting the reference for sinusoidal reference (\textit{bottom left}) and step reference (\textit{bottom right}), respectively.}
	\label{fig:SimSnapF}
\end{figure}

\subsection{Numerical Simulation}\label{simexp}
Before proceeding to conduct real-world experiments, we first verify the controller's behavior with a conceptual hybrid force/motion control task using the PyBullet simulation environment \cite{coumans2021}.
The QP problem that needs to be solved in the inner loop of the control architecture is handled by QP-solvers in python \cite{qpsolvers}.
The task is designed to be sliding the robot end-effector along a rigid board while tracking the given desired contact forces to mimic the real scoliosis test scenario as shown in the top row of Fig \ref{fig:SimSnapF}.
The robot end-effector frame $E$ and the base frame $\mathcal{I}$ are set the same as Fig. \ref{fig:exphuman}.  
The rigid board is set to be parallel to the $x-z$ plane of $\mathcal{I}$ with a distance of \SI{0.6}{\meter} away.
During performing the sliding task, the robot end-effector is required to be perpendicular to the board and the desired rotation matrix $\mathbf{R}_E^d$ is chosen as constant: 
$\mathbf{R}_E^d = \left[[0\; 0\; -1]^\T\;[0 \; 1 \; 0]^\T\;[1 \; 0 \; 0]^\T\right]$.
The start position of the robot end-effector is set to be $\begin{bmatrix}0.6 & 0 & 0.6\end{bmatrix}^\T \SI{}{\meter}$ and the end position is set to be
$\begin{bmatrix}0.6 & 0 & 0.9\end{bmatrix}^\T \SI{}{\meter}$.
We perform two similar tasks with the robot end-effector moving along a straight line connecting the start position and the end position at a constant speed of \SI{0.05}{\meter/\second}.
The robot end-effector is required to operate for \SI{6}{\second}, which results in a total sliding distance of \SI{0.3}{\meter} along the positive $z-$direction of the base frame. 
We let the end-effector track two different desired forces: a sinusoidal reference wave $f_1^d$ and a step reference $f_2^d$. 
Their designed numerical values are respectively given by
\begin{align*} 
f_1^d &= 5\sin(\pi t/3)+10\;\SI{}{\newton}, \quad t\in[0, 6]\; \SI{}{\second};\\ 
f_2^d &= \SI{15}{\newton}, \quad t\in[0, 6]\; \SI{}{\second}.
\end{align*}
The weights of the QP for different tasks of $\mathbf{v}$ as discussed in Section \ref{tasksspec} are selected to be $\begin{bmatrix} 0.1 & 1 & 1 & 1\end{bmatrix}^\T$ and the simulation step increments at a frequency of \SI{1}{\kilo\hertz}. 
It can be seen from the bottom row of Fig. \ref{fig:EXPContactF} that the tracking error $\epsilon$ between the measured and desired force remains relatively small in both cases with $\epsilon < \SI{0.01}{\newton}$, which validates the performance of the controller in simulation. 

\begin{figure}[t]
	\centering
	\includegraphics[width=\columnwidth]{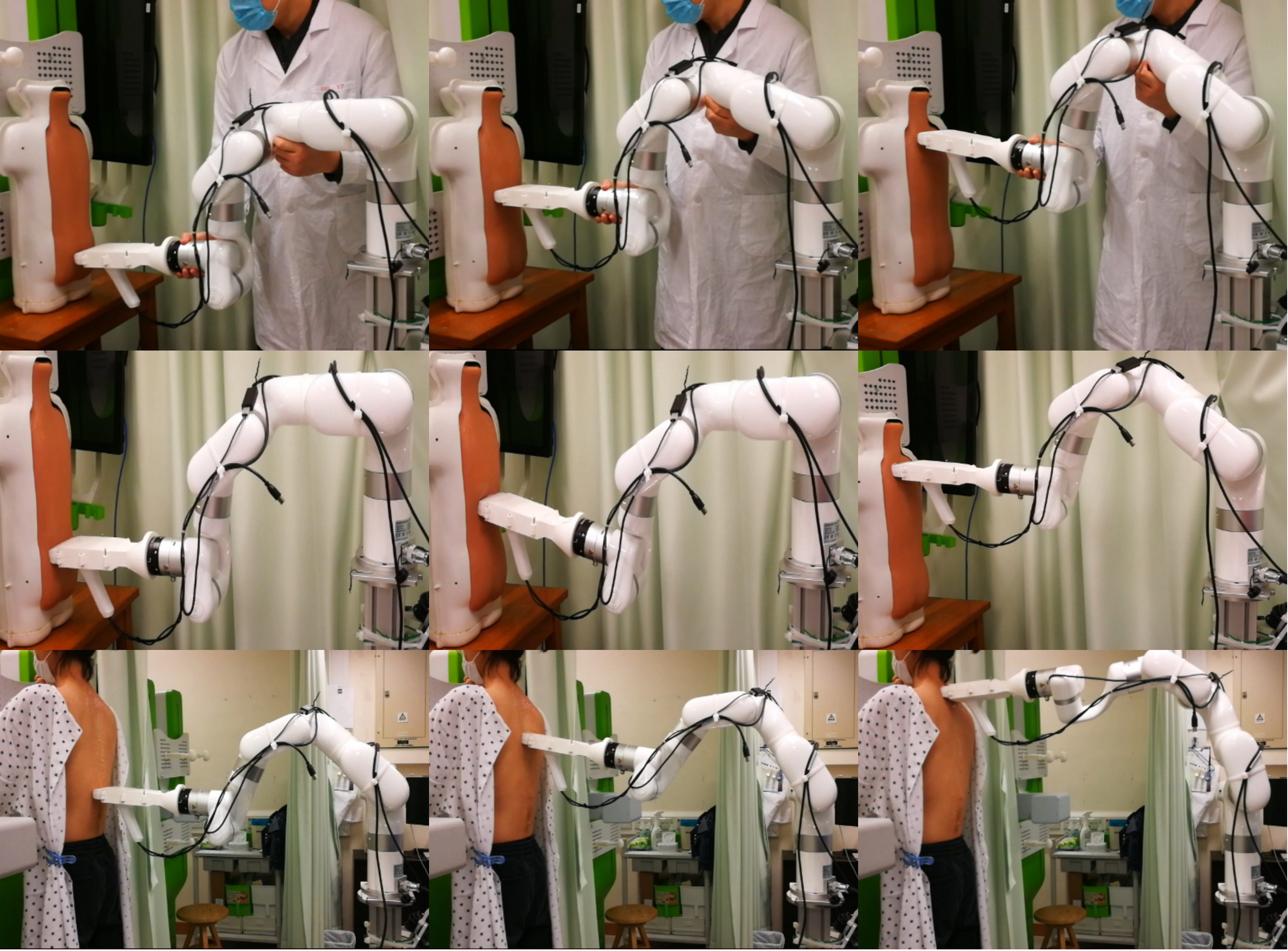}
	\caption{Snapshots of real experiment on scoliosis assessment demonstrated by a medical expert for impedance gains retrieval (\textit{top row}), execution on the phantom (\textit{middle row}), and execution on the human subject (\textit{bottom row}).}
	\label{fig:RealEXPDRH}
\end{figure}

\subsection{Experiments}\label{realexp}
The conducted experiments are about spinal image reconstruction with a phantom model and a human subject.
The robotic platform used throughout the experiments is an industrial robot manipulator called UFACTORY xArm 6, which is a fixed-base and serial robot manipulator having six DoFs. 
At the end effector, a USB ultrasound probe Sonoptek is mounted.
In addition, in order to enable direct force control, a six-axis Force/Torque sensor Robotiq FT300 is also installed at the robot end-effector.

The real experiments are conducted on the Scolioscan Air platform which is made up of a USB ultrasound probe and a tablet \cite{lai2021validation}.
The USB ultrasound probe captures ultrasound images at a frequency of \SI{7.5}{\mega\hertz} with a depth of \SI{6}{\centi\meter} and sends raw data at a frame rate of 10 $\mathrm{fps}$ to a desktop.
The aperture of the ultrasound probe is in the shape of a rectangular that has a length of \SI{80}{\milli\meter} and width \SI{15}{\milli\meter}.
The ultrasound images are organized in size of $640\times480$ pixels.
The tablet is responsible for receiving ultrasound images and coordinates from the ultrasound probe for 3D reconstruction of the scanned spine.
The robot manipulator xArm is connected via the TCP/IP protocol with the desktop.
We perform the experiments on the spinal phantom that contains a scoliosis spine inside as well as on a human volunteer subject; See the accompanying multimedia file.

\begin{figure}[t]
	\centering
	\includegraphics[width=1\columnwidth]{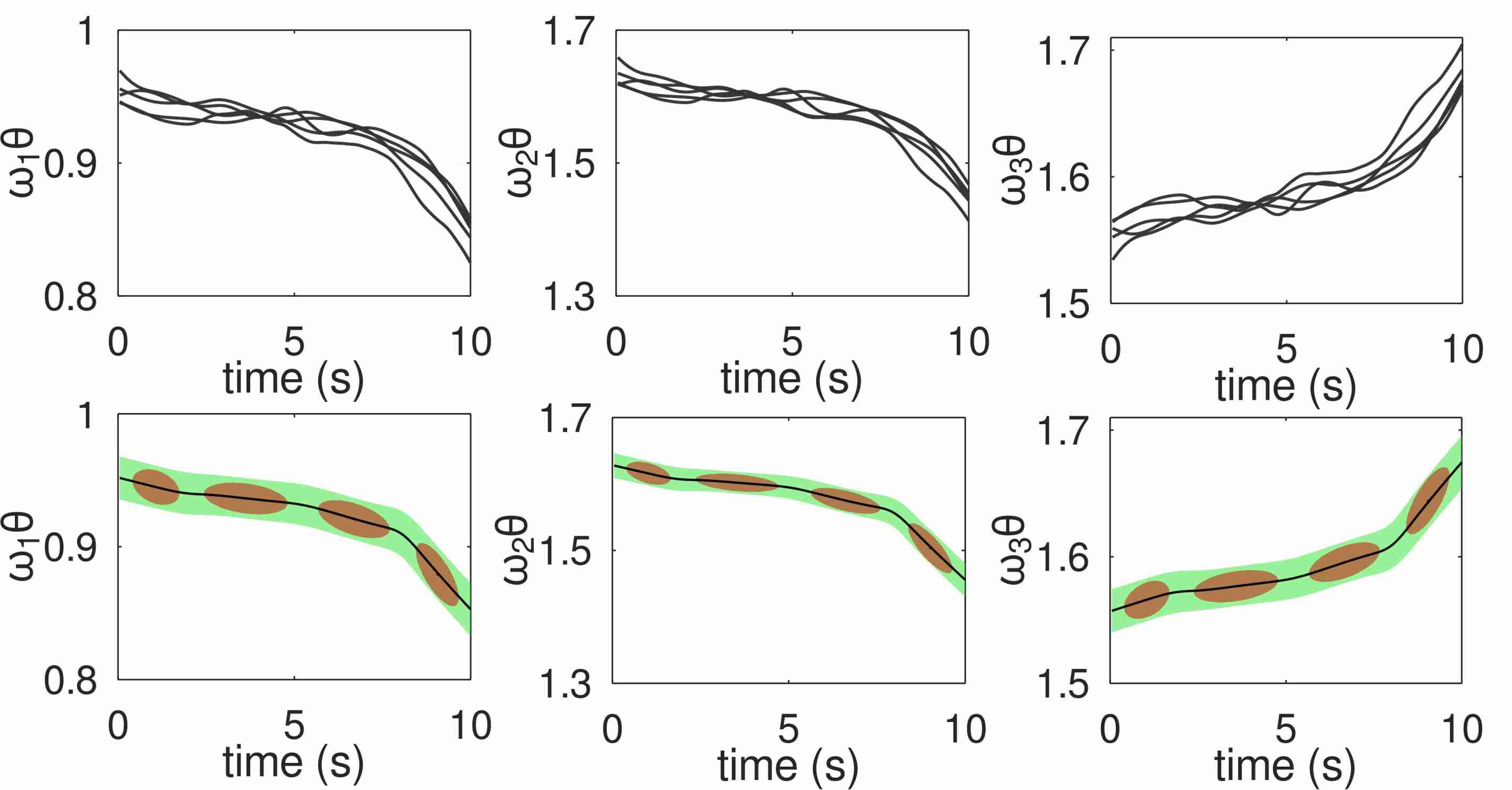}
	\caption{Multiple demonstrated trajectories of $\boldsymbol{\omega}\theta$ (\textit{top row}) and (\textit{bottom row}) the retrieved probabilistic trajectories with red ellipses denoting GMM components, green shallow area denoting the covariance and black line denoting the mean value.}
	\label{fig:omegathetaGMMGMR}
\end{figure}

\begin{figure}[t]
	\centering
	\includegraphics[width=1\columnwidth]{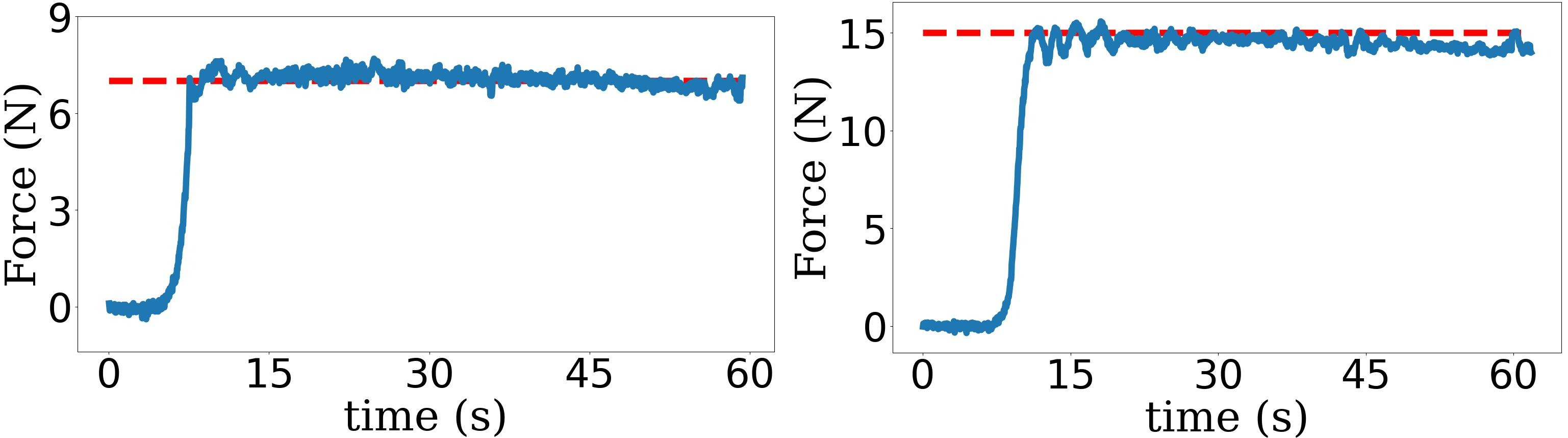}
	\caption{Tracking performance of the desired contact force on the phantom with red dashed line denoting the reference and blue line denoting measured value from the F/T sensor for $\SI{7}{\newton}$ (\textit{left}) and $\SI{15}{\newton}$ (\textit{right}), respectively.}
	\label{fig:EXPContactF}
\end{figure}

For estimating the impedance gains via employing the inverse of the covariance matrices, we provide multiple demonstrations on the phantom as shown in the top row of Fig. \ref{fig:RealEXPDRH}.
The demonstrated trajectories are conveyed via kinesthetic teaching by a medical expert.
The robot manipulator is set to be in the gravity-compensation mode such that it is light to drive. 
The back region, where the adipose tissue is thick, is expected to have more concentration of the demonstrated trajectories, implying that higher stiffness of the end-effector behavior would overcome the adipose's effects on the ultrasound image quality.    
Similarly, the back region with bumpy bones is expected to have more dispersion of the demonstrated trajectories, resulting in a lower stiffness of the end-effector in order to avoid injuring the patient.

In the experiment, the number of the demonstrated trajectories is determined as $M=5$. 
Here the input is chosen as temporal stamps.
During the human demonstration phase, the robot has a sampling period of \SI{0.05}{\second}.
The demonstration duration lasts for \SI{10}{\second}, which results in a total trajectory length of $N=200$ points.  
In the case of inconsistent demonstration duration, we simply employ some algorithms for aligning temporal sequences such as dynamic time warping.

For the processing of the collected dataset, we employ GMM/GMR for trajectory covariance retrieval as discussed in Section \ref{learnimp}.
The number of GMM components is chosen as $K=4$.
The parameters of the GMM, namely $\pi_k$, $\boldsymbol{\mu}_k$, and $\boldsymbol{\Sigma}_k$, are iteratively updated by means of the expectation maximization (EM) algorithm until convergence criteria is satisfied.
In the experiment, the iteration stopping criteria for defining convergence is set to be less than $1\mathrm{e}-4$ for the increase of the average log-likelihood value of EM in the current iteration.
Fig. \ref{fig:omegathetaGMMGMR} shows the modeling results of demonstrated orientation trajectories with GMM/GMR under the exponential parametrization using Octave. 
Probabilistic modeling of other demonstrated trajectories undergoes a similar procedure.

\begin{figure}[t]
	\centering
	\includegraphics[width=1\columnwidth]{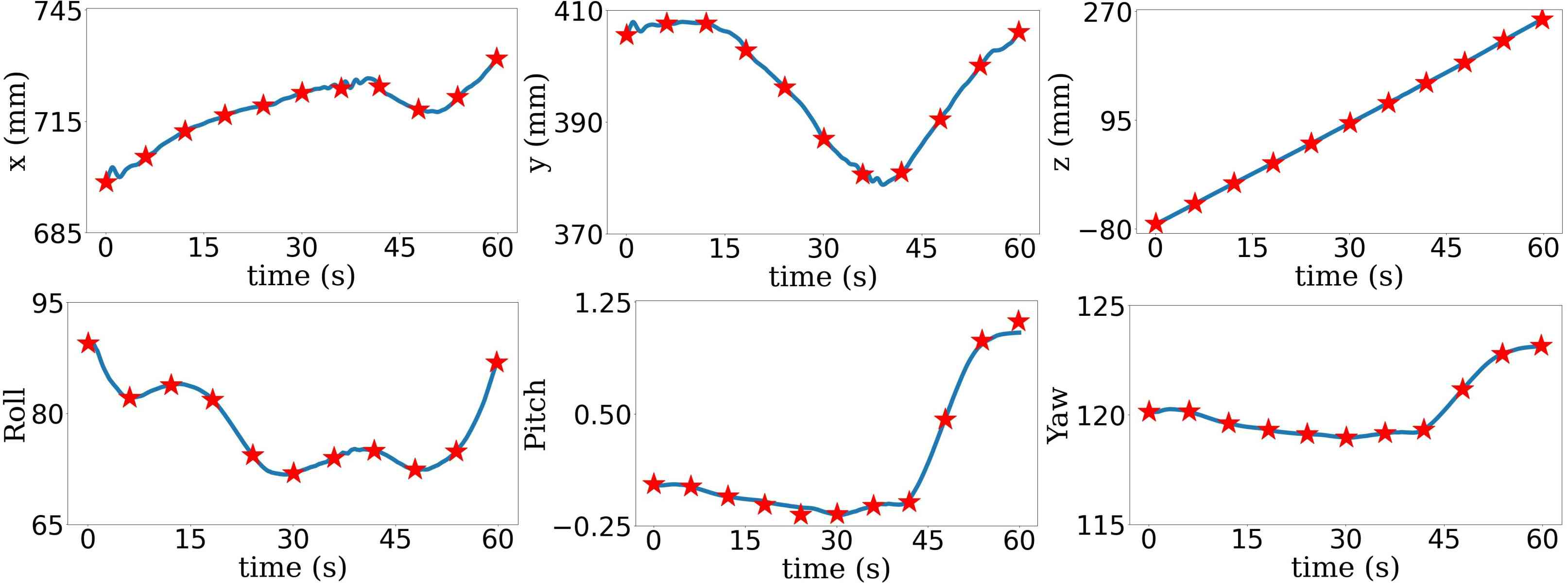}
	\caption{Tracking performance of end-effector position (\textit{top row}) and orientation (\textit{bottom row}) on the phantom with contact force being \SI{7}{\newton} where red stars denote the desired value.}
	\label{fig:RefMeasphantom7N}
\end{figure}

\begin{figure}[t]
	\centering
	\includegraphics[width=1\columnwidth]{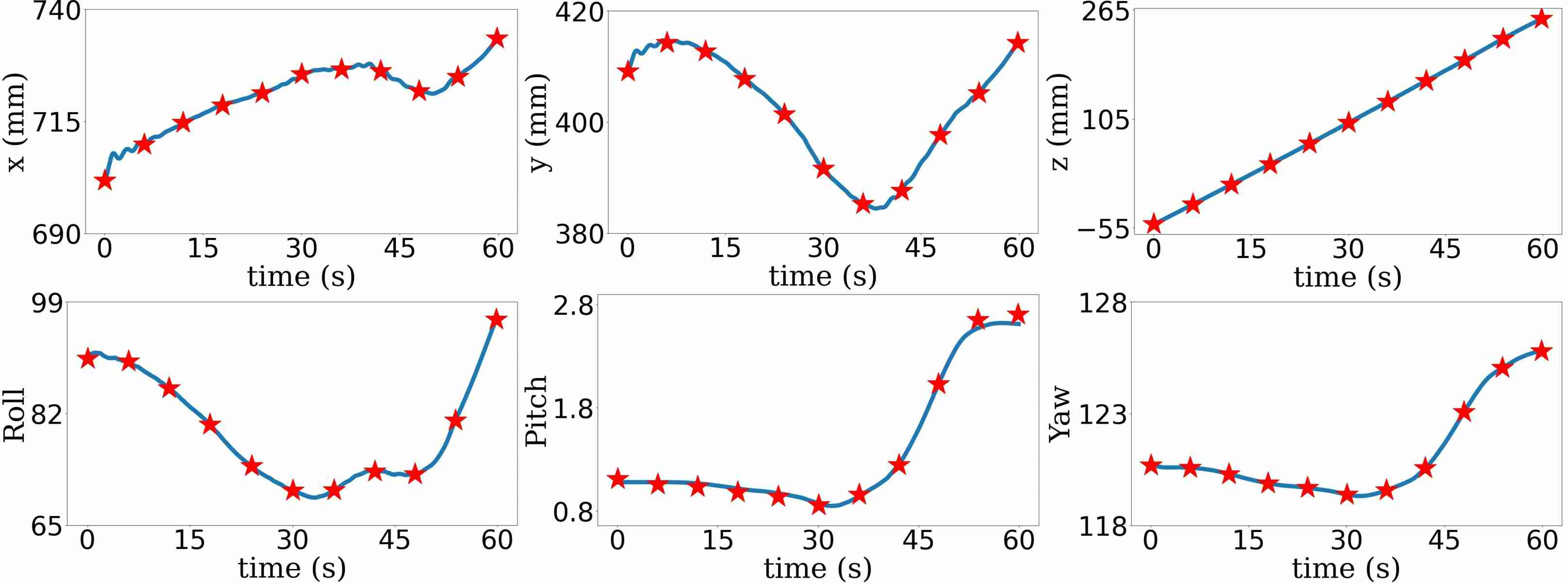}
	\caption{Tracking performance of end-effector position (\textit{top row}) and orientation (\textit{bottom row}) on the phantom with contact force being \SI{15}{\newton} where red stars denote the desired value.}
	\label{fig:RefMeasphantom15N}
\end{figure}

During the reproduction phase, our intention is to reconstruct the spines of the phantom and the human patient.
To do so, for both the phantom and the human subject, we first spread ultrasound gel over the backs as common practice in a clinic to avoid the air gap and enhance a tight contact between the back and the ultrasound transducer.
The ultrasound probe moves at a speed of \SI{0.003}{\meter/\second}.

\begin{figure}[t]
	\centering
	\includegraphics[width=0.9\columnwidth]{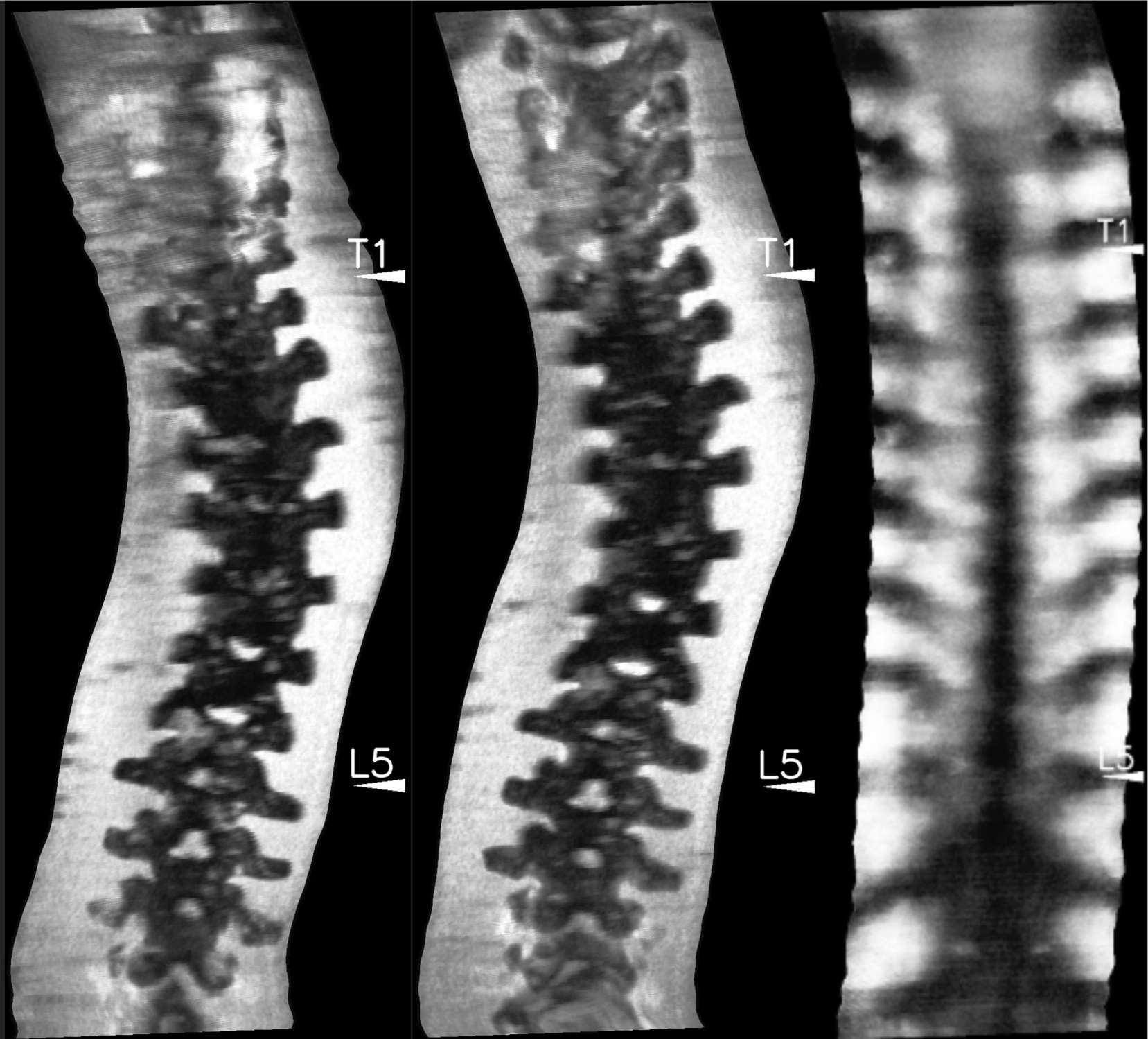}
	\caption{Spine reconstruction of a \emph{scoliosis} phantom model with $\SI{7}{\newton}$ (\textit{left}) and $\SI{15}{\newton}$ (\textit{middle}); Spine reconstruction of a human subject (no signs of scoliosis) with $\SI{10}{\newton}$ (\textit{right}).}
	\label{fig:SpineUSall}
\end{figure}

In the case of the phantom, two experiments are conducted with the desired contact force set to be \SI{7}{\newton} and \SI{15}{\newton}, respectively.
The procedure for scanning the spine of the phantom is shown in the middle row of Fig. \ref{fig:RealEXPDRH}.  
And Fig. \ref{fig:EXPContactF} shows the tracking performance of different contact forces.
It can be seen that the ultrasound probe can maintain a small force tracking error with its absolute value smaller than \SI{1}{\newton}, which is a typically allowable performance in practice.
The tracking performance of the end-effector pose expressed in the base frame is shown in Fig. \ref{fig:RefMeasphantom7N} and Fig. \ref{fig:RefMeasphantom15N}.
The obtained spine image of the phantom is shown in the left and middle column of Fig. \ref{fig:SpineUSall}.
It is reasonable to observe that the case with the bigger contact force has a more clear spine image.

\begin{figure}[t]
	\centering
	\includegraphics[width=1\columnwidth]{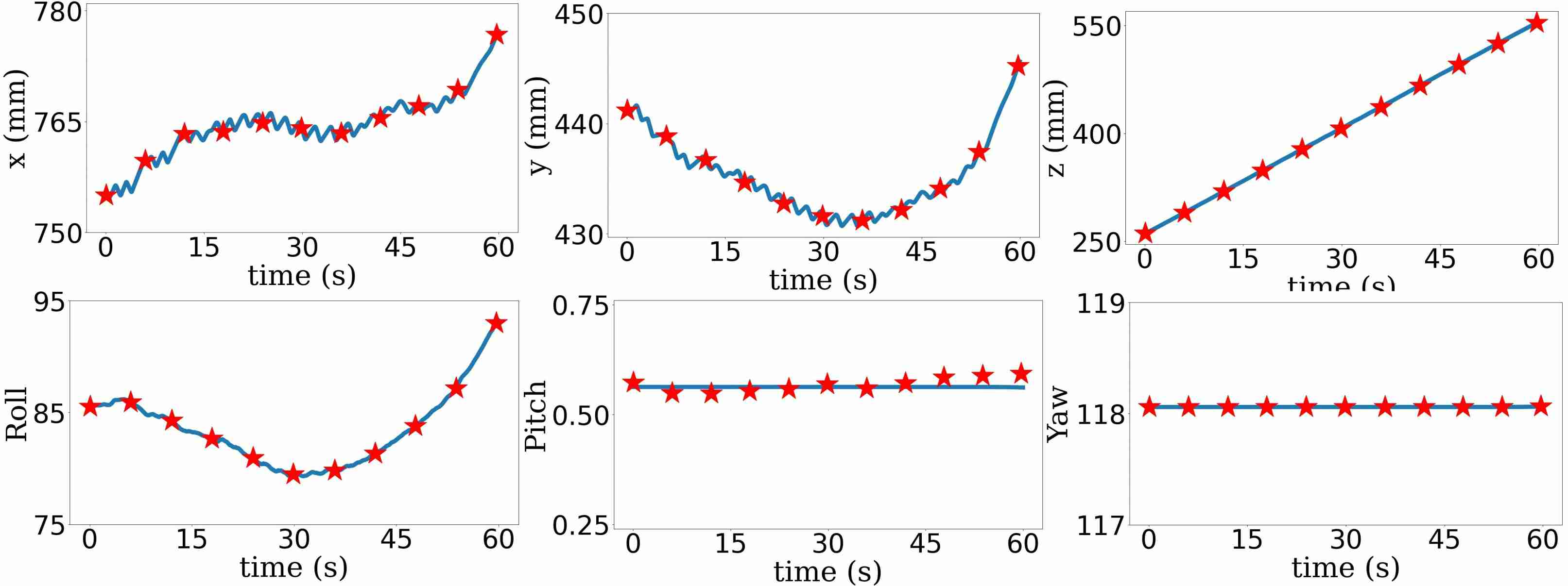}
	\caption{Tracking performance of end-effector position (\textit{top row}) and orientation (\textit{bottom row}) on the human subject with contact force being \SI{10}{\newton} where red stars denote the desired value.}
	\label{fig:RefMeashuman10N}
\end{figure}

In the case of testing with a human subject, the desired contact force usually depends on the Body Mass Index (BMI).
Here we empirically set it to be \SI{10}{\newton}.
The procedure for scanning the spine of the human subject is shown in the bottom row of Fig. \ref{fig:RealEXPDRH}. 
The corresponding tracking performance of the end-effector pose expressed in the base frame is shown in Fig. \ref{fig:RefMeashuman10N}, where is error is negligible in terms of our application.
And the obtained spine image of the human subject is shown in the right column of Fig \ref{fig:SpineUSall}.
The quality of the image is clear enough for medical personnel to measure the curvature of the volunteer's spine.

	\section{Discussions and Conclusion}\label{conclusion}
In this paper, we presented a control architecture for autonomous scoliosis assessment with a robotic manipulator.
The proposed control architecture is composed of two loops, an outer loop that outputs the virtual control signal and an inner loop that tracks the virtual control signal.
Compared with designing control laws analytically, the employed optimization-based control strategy is easier to specify.
Furthermore, it has been shown that the formulated optimization problem is in fact a form of QP, which can be solved very fast. 
Compared with reinforcement learning-based motion generation, our approach offers principled control synthesis and presents a more reliable and explainable behavior for the robot.
Also, the convergence issue would prohibit reinforcement learning from deployment in the real world \cite{li2020lambda}.
And such concern will be more severe under the medical treatment context.


Regarding the gains profile for impedance regulation, we resorted to the general framework of learning by demonstrations where the inverse of the retrieved covariance matrices were set as the desired impedance gains.
For learning stiffness of the rotation matrices, we proposed to re-parameterize the rotation matrix with the exponential coordinate such that naive GMM/GMR could be applied directly without the need of extending it to manifold-structured data. 

The effectiveness of the proposed approach is verified with both simulation and real experiments.
Specifically, we successfully applied our approach to a phantom and a human subject for scoliosis assessment.
Although the proposed approach can function well for the spine scanning with ultrasound, certainly there is still room for future work to improve the platform for real deployment.
For example, now it still requires a medical assistant to spread the ultrasound gel for the usage of the platform.
This procedure could also be automated by the robot manipulator to increase autonomy level.
Also, to locate the position of the spinous process more precisely, other forms of sensory information like thermal images could be incorporated for sensor fusion in addition to ultrasound images \cite{hu2021radiation}.

	
	\bibliographystyle{IEEEtran}
	\bibliography{bibliography}
	
	
	\vfill
	
\end{document}